\documentclass[sigconf,authorversion,nonacm]{acmart}

\AtBeginDocument{%
  \providecommand\BibTeX{{%
    \normalfont B\kern-0.5em{\scshape i\kern-0.25em b}\kern-0.8em\TeX}}}

\setcopyright{acmcopyright}
\copyrightyear{2023}
\acmYear{2023}

\acmConference[CUI2023]{ACM Conversational User Interfaces}{July 19--21,
  2023}{Eindhoven, The Netherlands}
\acmPrice{15.00}
\acmISBN{978-1-4503-XXXX-X/18/06}

\begin{document}

\title{Utilization of Non-verbal Behaviour and Social Gaze in Classroom Human-Robot Interaction Communications}

\author{Sahand Shaghaghi\ \ \ Pourya Aliasghari\ \ \ Bryan Tripp\ \ \ Kerstin Dautenhahn\ \ \ Chrystopher Nehaniv}
\email{s2shagha@uwaterloo.ca}
\orcid{0000-0002-4666-7447}




\affiliation{%
  \institution{University of Waterloo}
  \city{Waterloo}
  \state{Ontario}
  \country{Canada}
}

\renewcommand{\shortauthors}{Shaghaghi, et al.}

\begin{abstract}
This abstract explores classroom Human-Robot Interaction (HRI) scenarios with an emphasis on the adaptation of human-inspired social gaze models in robot cognitive architecture to facilitate a more seamless social interaction. First, we detail the HRI scenarios explored by us in our studies followed by a description of the social gaze model utilized for our research. We highlight the advantages of utilizing such an attentional model in classroom HRI scenarios. We also detail the intended goals of our upcoming study involving this social gaze model. 
\end{abstract}

\ccsdesc[500]{Human-centered computing~Empirical studies in interaction design}
\ccsdesc[300]{Human-centered computing~Empirical studies in HCI}
\keywords{Human-Robot Interaction, Social Interactions, Gaze, Visual Perception, Language Learning, Classroom Interactions}



\maketitle

\section{Introduction}
The use of robots in the classroom is of interest to educators and researchers alike. However, utilization of robots in the classroom is presently under investigation. The majority of the humanoid robotic platforms aimed at classroom interactions are presently not capable of long-term engagement with participants. These robots face frequent failures~\cite{honig_understanding_2018} and on occasion will fail in conversational settings and are unable to recover from these failures~\cite{kontogiorgos_embodiment_2020}. Robot failures could be categorized into technical failures and interaction failures~\cite{honig_understanding_2018}. The interaction failures are in part related to social norm violations or human errors~\cite{honig_understanding_2018}, communication failures, and misalignment failures. These interaction failures are either detectable by the interaction partner or purely internal and non-detectable~\cite{honig_understanding_2018}.

Communication failures and misalignment failures both arise from the systemic interaction process. Miscommunication is a multi-person phenomenon composed of the inter-relation involving produced actions and content and the reaction to these actions and content~\cite{healey_editors_2018}. Miscommunication amounts to more than noise and includes “misarticulations, malapropisms, use of a 'wrong' word, unavailability of a word when needed, and trouble on the part of the recipient in understanding”~\cite{healey_editors_2018,schegloff_preference_1977}. Conversational analysis has demonstrated that repair relating to these miscommunications is incremental and jointly managed~\cite{byun_first_2018, mccabe_miscommunication_2018, purver_computational_2018}, in part universal across languages~\cite{dingemanse_is_2013, dingemanse_universal_2015} and also, in part generalizable across modalities including visual communication~\cite{manrique_suspending_2015, mortensen_visual_2012}.

Misalignment failures happen due to failure in “knowing” the intended goal of a communicated joint action by the receiver. For example, such failure could arise when the robot is asked to pay attention to a box on the table followed by inaction by the robot since it does not know what a box is. Various failures can be mitigated or addressed by the enhancement of {\it mutual gearing}~\cite{cowley_distributed_2005, cowley_distributed_2007} into robotic communication design.

In our research, we are interested in how social gaze could assist with deterrence and recovery from these failures. Failures are not always detectable, especially by robotic agents. Even if the system could detect the failure, it might not be able to classify the type of failure or have any mechanism to remedy the failure. There are various possible failure recovery methods for the robot including justification, apologies, compensation, and options for the user~\cite{Correia2018ExploringTrack,honig_understanding_2018,10.1145/3472307.3484184,Fratczak2021RobotInteraction,Lee2010GracefullyServices}. Incorporation of social gaze into robotic cognitive architecture is beneficial since it could engage the partner and the robot through joint attention using the visual stimuli present surrounding the interaction. The presence of such joint attention is valuable when failure warnings are needed. Heuristic social gaze behaviour of the robot could provide the human partner with better knowledge of the interaction and more predictability of possible upcoming failures of system-level interaction with the robot. Social gaze as an additional channel in conjunction with non-verbal behaviour could sustain the interaction if there are failures in other modes of communication (e.g. speech). Gaze cues could be used to communicate failure (e.g. withdrawn eye contact), and to mitigate failure disruptiveness~\cite{booth_errors_1991}.

The more sensory channels available to the robot through its cognitive architecture, there may be better the chances of the robot communicating effectively in social scenarios~\cite{eldardeer_biological_2021} and the better may be the chances of the robot recovering from communication failures. One of the present goals in robotics is the creation of attentional mechanisms for robots which could help ground them in social conversational scenarios. A major component of such an attentional mechanism is visual. The present research conducted by us explores the implementation of a social gaze model rooted in visual psychology research~\cite{jording_social_2018} for the iCub robot~\cite{Metta2008TheCognition}. This abstract partly covers the steps taken by us for this implementation. The goal of our present line of research is to make HRI more natural, smooth, and robust.

This abstract details two experimental studies. The first study completed by us revolves around classroom conversational HRI utilizing pointing and deterministic gaze by the iCub robot. This study focused on participants observing language learning in human-robot teaching scenarios and the perception of this interaction by third-person observer participants. The study design makes use of the Esperanto language and object name learning by the observer participants. Teaching efficacy is measured for all conditions in this study. For our upcoming second study, we are incorporating human-inspired social gaze as an attentional mechanism into the robotic cognitive architecture. We are interested in seeing how the incorporation of social gaze could impact the overall quality of interaction and participant perception of the robot.

\section{Classroom HRI}
Classroom HRI refers to human-robot interactions which take place in a formal teaching context in which the robot is either a teacher, a student or a collaborator. These interactions usually take place in a classroom setting. Humanoid robotic platforms are being considered for use in classroom settings as teacher aids. These robotic platforms show promise in relation to educational settings due to the overall positive perception of these robots by students~\cite{tanaka_children_2012}. Robots have been studied in the capacities of teacher, student and collaborator with frequent focus on language learning tasks~\cite{gavrilova_pilot_2019}. 

The robotic platforms being experimented on in classroom settings are not able to operate in unstructured scenarios. This means that these robots are not able to carry on meaningful conversations when the interactions fall beyond what the robot is tasked with. This could be a shortfall for conversational situations involving loosely specified tasks. These robots are also not well situated to recover from physical failures. As an example, the iCub robot would need at least a few minutes to recover from a major failure (system requiring a reboot) which will affect the quality of interaction and the participant’s trust in the robot.

Both studies conducted by us make use of the iCub robot in classroom settings. In the scenarios involved in these studies, the robot engages in active language learning with an actor. In these studies, the robot takes on different roles and personality traits. 

\section{First Study Design}
This study, which is also the foundation for our second upcoming study, engages the iCub robot with the actor in language learning scenarios involving learning the names of specific shapes in the Esperanto language. The robot takes on different personality traits and roles in a series of teaching scenarios. This study has a two-by-three design in which the robot takes on two personality types and three roles. The personalities embodied by the robot are introverted and extroverted and the roles embodied by the robot are teacher, student, and collaborator. In these series of teaching scenarios names of shapes are taught by the teacher who then evaluates student’s learning abilities using recall. In this study, we were mainly interested in observing how the robot's role and personality impact observer participants’ perception of the robot. We were also interested in teaching efficacy associated with each of the teaching scenarios included in this study.

This study explores non-verbal cues (i.e. posture, gaze, voice) expressed by the robot in communication. Non-verbal behaviour associated with different personalities affect the quality of the interaction differently and could enhance the quality of interaction in specific scenarios, for specific populations.

\subsection{iCub Robot Utilization: Expressing Extrovert vs. Introvert Traits}
The iCub humanoid robot is designed for researching embodied cognition and conducting HRI experiments~\cite{Metta2008TheCognition}. To conduct our experiments, the robot needed to exhibit effective verbal and nonverbal behaviour as a teacher, student, or collaborator in different scenarios. These behaviour included speaking, gazing, and pointing at objects. Six pre-defined scripts were created, one for each scenario, with the robot dialogue remaining consistent across scenarios to avoid bias. The robot's personality was altered by varying its verbal and nonverbal behaviour, and its role was changed by incorporating different conversations and turn-taking strategies. 

Regarding verbal aspects, the robot's speech was generated using the built-in text-to-speech module, simulating lip movements by changing the shape of the robot's mouth. The extroverted robot spoke faster with a higher pitch, achieved by modifying the text-to-speech properties. Arm movements were programmed to accompany the robot's speech, making it appear more natural. The extroverted robot used expanded hand postures with bent elbows, while the introverted robot maintained a straight arm configuration close to its body. 

Pointing behaviour was pre-defined by closing all fingers except the index finger and positioning the robot's arm accordingly for referencing. The extroverted robot bent its hip to point from a closer distance, while the introverted robot stood straight and used only its arm for pointing. When pointing, the robot also gazed at the object, but if it needed to speak while pointing, it gazed at the interaction partner (actor). When the human partner (actor) pointed to an object, the robot gazed at the object to appear attentive. 

\section{Social Gaze Space Theory and Second Study Design}
The second study, taking place this autumn, aims to incorporate social gaze models into a robot cognitive architecture as an attentional mechanism. We utilize the Social Gaze Space model (SGS)~\cite{jording_social_2018} rooted in human psychology literature. SGS is a heuristic model of gaze. This model investigates gaze interactions involving two humans and few objects. This model defines five distinct states of gaze for participants: Partner-Oriented (PO), Object-Oriented (OO), Introspective (INT), Responding Joint Attention (RJA), and Initiating Joint Attention (IJA)~\cite{jording_social_2018}. Below the five gaze states are detailed:

\begin{enumerate}
    \item Partner-Oriented (PO): In this state, participant 1 exclusively pays attention to participant 2. This exclusive attentional state leads to participant 2 eventually focusing on participant 1’s eyes, which reveals participant 1’s emotional and attentional state.
    \item Object-Oriented (OO): In this state, participant 1’s attention is entirely focused on the objects present in the environment and not participant 2. This state is different from joint attention states where participant 1 splits their attention between objects and participant 2.
    \item Introspective (INT): In this state, participant 1 refrains from paying attention to the objects and participant 2 and instead dwells on their own inner state. 
    \item Responding Joint Attention (RJA): In this state, participant 1 follows participant 2’s initiation and leading of gaze interaction. For example, in RJA, participant 2 would first look participant 1 in the eyes and then look at an object, which would lead participant 1 to look at the same object.
    \item Initiating Joint Attention (IJA): In this state, participant 1 initiates and leads the gaze interaction with participant 2. In IJA, participant 1 would take the lead in looking participant 2 in the eyes and then gazing at an object, leading participant 2 to look at that object. The main difference between this gaze state (IJA) and the previous gaze state (RJA) is that here participant 1 is taking the lead role while in the previous state, participant 1 takes a follower role.
\end{enumerate}

Depending on each participant’s gaze state, the allocation of the overall gaze state in the model could be determined. A gaze state of the two-partner system is given by an ordered pair of the above states in which participant 1 represents the self and participant 2 represents the other. In this model gaze state allocation is either stable or unstable. The system has a tendency to move towards stability. The state PO/PO acts as a gate which could lead to more mutually interactive states.

\subsection{Robotic System Implementation}
In order to design an attentional architecture which complies with the SGS model, we took advantage of a number of vision libraries created in the past:

\begin{itemize}
    \item Ptgaze, a Python library which makes it possible for the iCub robot to perceive the participants' gaze direction.~\cite{hysts_pytorch_mpiigaze_demo_2022}. We are also using this library to locate the participants’ face location in the visual frame.
    \item ETHgaze, a deep neural model employed by Ptgaze library for participant gaze direction detection~\cite{zhang_eth-xgaze_2020}.
    \item YOLO, a deep neural model utilized to identify and locate objects in the visual frame~\cite{jocher_yolov5_2021}. 
    \item The depth perception module for the iCub robot to calculate participant and object distances from the robot~\cite{pasquale_enabling_2016}.

\end{itemize}

We also developed a decision processes algorithm. This algorithm allows the adaptation of the SGS model for the iCub robotic platform. This algorithm defines a five-by-five state space model with the state transitions being a hybrid of rule-based and probabilistic in accordance with the SGS model details.

This implementation utilizes the iCub’s onboard anthropomorphic vision system without assistance from other external camera feeds (unlike previous implementations which make use of external Kinect cameras~\cite{lohan_tutor_2012}). This implementation is novel and adds to the already ground-breaking research in this field on two fronts: First, utilization of the internal-only camera system of the iCub robot which is much more biomimetically plausible. Second, the implementation of such an architecture for an onboard anthropomorphic vision system rather than a fix-mounted stereo system.

\section{Intended Goals of the Planned Upcoming Study}
Our upcoming study aims to assess participants' perception of the robot and the quality of interaction when a social gaze model is incorporated into robot’s cognitive architecture. This study utilizes the iCub robot and will involve the participants in an in-person capacity. We are aiming to use various questionnaires to assess the participants' perception of the robot and the quality of interaction. This includes ROSAS \cite{carpinella_robotic_2017}, TIPI \cite{gosling_very_2003}, and questionnaires to measure the social acceptability of the robot. We aim to investigate the following research questions:

\begin{enumerate}
    \item Could heuristic and cognitive models of non-verbal interaction be used to achieve communicative social gaze in robots in order to influence:
    \begin{itemize}
    \item Participants’ perception of the robot.
    \item Participants’ expectations, experience of interaction and “Quality of interaction”.
    \item Efficacy of interaction.
    \end{itemize}
    \item How would participant’s and robot’s taking on of different roles (i.e. teacher, students and
collaborator), and the robot’s non-verbal expressive behaviour affect the social interactions
in the context of experiments involving the aforementioned non-verbal interaction model
investigations?
\item How could one construct a cognitive architecture utilizing gaze state transitions as prescribed by SGS which could modulate interactive gaze behaviour by robots in dyadic interactions?
\item Does participants’ perception/view of the robot (e.g. in terms of likability, intelligence,
etc.) differ depending on robot’s gaze behaviour?
\item Are participants paying attention to and learning about the task that is being taught?
\end{enumerate}

\section{Conclusion}
Humans use a variety of non-verbal cues that can mitigate failures of verbal communication. Our studies model cues related to body language and mutual gaze, and assess their influence on human-robot interactions, including how effectively a robot can teach new information to a human.

\section*{Acknowledgments}
This research was undertaken, in part, thanks to funding from the Canada 150 Research Chairs Program.


\bibliographystyle{ACM-Reference-Format}
\bibliography{references,new-refs}


\begin{thebibliography}{30}


\ifx \showCODEN    \undefined \def \showCODEN     #1{\unskip}     \fi
\ifx \showDOI      \undefined \def \showDOI       #1{#1}\fi
\ifx \showISBNx    \undefined \def \showISBNx     #1{\unskip}     \fi
\ifx \showISBNxiii \undefined \def \showISBNxiii  #1{\unskip}     \fi
\ifx \showISSN     \undefined \def \showISSN      #1{\unskip}     \fi
\ifx \showLCCN     \undefined \def \showLCCN      #1{\unskip}     \fi
\ifx \shownote     \undefined \def \shownote      #1{#1}          \fi
\ifx \showarticletitle \undefined \def \showarticletitle #1{#1}   \fi
\ifx \showURL      \undefined \def \showURL       {\relax}        \fi
\providecommand\bibfield[2]{#2}
\providecommand\bibinfo[2]{#2}
\providecommand\natexlab[1]{#1}
\providecommand\showeprint[2][]{arXiv:#2}

\bibitem[Aliasghari et~al\mbox{.}(2021)]%
        {10.1145/3472307.3484184}
\bibfield{author}{\bibinfo{person}{Pourya Aliasghari}, \bibinfo{person}{Moojan
  Ghafurian}, \bibinfo{person}{Chrystopher~L. Nehaniv}, {and}
  \bibinfo{person}{Kerstin Dautenhahn}.} \bibinfo{year}{2021}\natexlab{}.
\newblock \showarticletitle{How Do Different Modes of Verbal Expressiveness of
  a Student Robot Making Errors Impact Human Teachers’ Intention to Use the
  Robot?}. In \bibinfo{booktitle}{\emph{Proceedings of the 9th International
  Conference on Human-Agent Interaction}} (Virtual Event, Japan)
  \emph{(\bibinfo{series}{HAI '21})}. \bibinfo{publisher}{Association for
  Computing Machinery}, \bibinfo{address}{New York, NY, USA},
  \bibinfo{pages}{21–30}.
\newblock
\showISBNx{9781450386203}
\urldef\tempurl%
\url{https://doi.org/10.1145/3472307.3484184}
\showDOI{\tempurl}


\bibitem[Booth(1991)]%
        {booth_errors_1991}
\bibfield{author}{\bibinfo{person}{Paul~A Booth}.}
  \bibinfo{year}{1991}\natexlab{}.
\newblock \showarticletitle{Errors and theory in human-computer interaction}.
\newblock \bibinfo{journal}{\emph{Acta Psychologica}} \bibinfo{volume}{78},
  \bibinfo{number}{1-3} (\bibinfo{year}{1991}), \bibinfo{pages}{69--96}.
\newblock
\newblock
\shownote{Publisher: Elsevier}.


\bibitem[Byun et~al\mbox{.}(2018)]%
        {byun_first_2018}
\bibfield{author}{\bibinfo{person}{Kang-Suk Byun}, \bibinfo{person}{Connie de
  Vos}, \bibinfo{person}{Anastasia Bradford}, \bibinfo{person}{Ulrike Zeshan},
  {and} \bibinfo{person}{Stephen~C. Levinson}.}
  \bibinfo{year}{2018}\natexlab{}.
\newblock \showarticletitle{First {Encounters}: {Repair} {Sequences} in
  {Cross}-{Signing}}.
\newblock \bibinfo{journal}{\emph{Topics in Cognitive Science}}
  \bibinfo{volume}{10}, \bibinfo{number}{2} (\bibinfo{year}{2018}),
  \bibinfo{pages}{314--334}.
\newblock
\urldef\tempurl%
\url{https://doi.org/10.1111/tops.12303}
\showDOI{\tempurl}
\newblock
\shownote{\_eprint:
  https://onlinelibrary.wiley.com/doi/pdf/10.1111/tops.12303}.


\bibitem[Carpinella et~al\mbox{.}(2017)]%
        {carpinella_robotic_2017}
\bibfield{author}{\bibinfo{person}{Colleen~M. Carpinella},
  \bibinfo{person}{Alisa~B. Wyman}, \bibinfo{person}{Michael~A. Perez}, {and}
  \bibinfo{person}{Steven~J. Stroessner}.} \bibinfo{year}{2017}\natexlab{}.
\newblock \showarticletitle{The {Robotic} {Social} {Attributes} {Scale}
  ({RoSAS}): {Development} and {Validation}}. In \bibinfo{booktitle}{\emph{2017
  12th {ACM}/{IEEE} {International} {Conference} on {Human}-{Robot}
  {Interaction} ({HRI}}}. \bibinfo{pages}{254--262}.
\newblock


\bibitem[Correia et~al\mbox{.}(2018)]%
        {Correia2018ExploringTrack}
\bibfield{author}{\bibinfo{person}{Filipa Correia}, \bibinfo{person}{Carla
  Guerra}, \bibinfo{person}{Samuel Mascarenhas}, \bibinfo{person}{Francisco~S.
  Melo}, {and} \bibinfo{person}{Ana Paiva}.} \bibinfo{year}{2018}\natexlab{}.
\newblock \showarticletitle{Exploring the Impact of Fault Justification in
  Human-Robot Trust}. In \bibinfo{booktitle}{\emph{Proceedings of the 17th
  International Conference on Autonomous Agents and MultiAgent Systems}}
  \emph{(\bibinfo{series}{AAMAS '18})}. \bibinfo{publisher}{International
  Foundation for Autonomous Agents and Multiagent Systems},
  \bibinfo{pages}{507–513}.
\newblock
\urldef\tempurl%
\url{https://doi.org/10.5555/3237383.32374598}
\showDOI{\tempurl}


\bibitem[Cowley(2005)]%
        {cowley_distributed_2005}
\bibfield{author}{\bibinfo{person}{Stephen~J Cowley}.}
  \bibinfo{year}{2005}\natexlab{}.
\newblock \showarticletitle{A distributed view of language origins}. In
  \bibinfo{booktitle}{\emph{Second {International} {Symposium} on the
  {Emergence} and {Evolution} of {Linguistic} {Communication} ({EELC}'05).
  {Hatfield}, {UK}: {AISB} {Press}}}. \bibinfo{publisher}{Citeseer},
  \bibinfo{pages}{23--27}.
\newblock


\bibitem[Cowley(2007)]%
        {cowley_distributed_2007}
\bibfield{author}{\bibinfo{person}{Stephen~J. Cowley}.}
  \bibinfo{year}{2007}\natexlab{}.
\newblock \showarticletitle{Distributed {Language}: {Biomechanics},
  {Functions}, and the {Origins} of {Talk}}.
\newblock In \bibinfo{booktitle}{\emph{Emergence of {Communication} and
  {Language}}}, \bibfield{editor}{\bibinfo{person}{Caroline Lyon},
  \bibinfo{person}{Chrystopher~L. Nehaniv}, {and} \bibinfo{person}{Angelo
  Cangelosi}} (Eds.). \bibinfo{publisher}{Springer London},
  \bibinfo{address}{London}, \bibinfo{pages}{105--127}.
\newblock
\showISBNx{978-1-84628-779-4}
\urldef\tempurl%
\url{https://doi.org/10.1007/978-1-84628-779-4_6}
\showDOI{\tempurl}


\bibitem[Dingemanse et~al\mbox{.}(2015)]%
        {dingemanse_universal_2015}
\bibfield{author}{\bibinfo{person}{Mark Dingemanse}, \bibinfo{person}{Seán~G.
  Roberts}, \bibinfo{person}{Julija Baranova}, \bibinfo{person}{Joe Blythe},
  \bibinfo{person}{Paul Drew}, \bibinfo{person}{Simeon Floyd},
  \bibinfo{person}{Rosa~S. Gisladottir}, \bibinfo{person}{Kobin~H. Kendrick},
  \bibinfo{person}{Stephen~C. Levinson}, \bibinfo{person}{Elizabeth Manrique},
  \bibinfo{person}{Giovanni Rossi}, {and} \bibinfo{person}{N.~J. Enfield}.}
  \bibinfo{year}{2015}\natexlab{}.
\newblock \showarticletitle{Universal {Principles} in the {Repair} of
  {Communication} {Problems}}.
\newblock \bibinfo{journal}{\emph{PLOS ONE}} \bibinfo{volume}{10},
  \bibinfo{number}{9} (\bibinfo{date}{Sept.} \bibinfo{year}{2015}),
  \bibinfo{pages}{1--15}.
\newblock
\urldef\tempurl%
\url{https://doi.org/10.1371/journal.pone.0136100}
\showDOI{\tempurl}
\newblock
\shownote{Publisher: Public Library of Science}.


\bibitem[Dingemanse et~al\mbox{.}(2013)]%
        {dingemanse_is_2013}
\bibfield{author}{\bibinfo{person}{Mark Dingemanse}, \bibinfo{person}{Francisco
  Torreira}, {and} \bibinfo{person}{N.~J. Enfield}.}
  \bibinfo{year}{2013}\natexlab{}.
\newblock \showarticletitle{Is “{Huh}?” a {Universal} {Word}?
  {Conversational} {Infrastructure} and the {Convergent} {Evolution} of
  {Linguistic} {Items}}.
\newblock \bibinfo{journal}{\emph{PLOS ONE}} \bibinfo{volume}{8},
  \bibinfo{number}{11} (\bibinfo{date}{Nov.} \bibinfo{year}{2013}),
  \bibinfo{pages}{1--10}.
\newblock
\urldef\tempurl%
\url{https://doi.org/10.1371/journal.pone.0078273}
\showDOI{\tempurl}
\newblock
\shownote{Publisher: Public Library of Science}.


\bibitem[Eldardeer et~al\mbox{.}(2021)]%
        {eldardeer_biological_2021}
\bibfield{author}{\bibinfo{person}{Omar Eldardeer}, \bibinfo{person}{Jonas
  Gonzalez-Billandon}, \bibinfo{person}{Lukas Grasse}, \bibinfo{person}{Matthew
  Tata}, {and} \bibinfo{person}{Francesco Rea}.}
  \bibinfo{year}{2021}\natexlab{}.
\newblock \showarticletitle{A {Biological} {Inspired} {Cognitive} {Framework}
  for {Memory}-{Based} {Multi}-{Sensory} {Joint} {Attention} in {Human}-{Robot}
  {Interactive} {Tasks}}.
\newblock \bibinfo{journal}{\emph{Frontiers in Neurorobotics}}
  \bibinfo{volume}{15} (\bibinfo{year}{2021}).
\newblock
\showISSN{1662-5218}
\urldef\tempurl%
\url{https://doi.org/10.3389/fnbot.2021.648595}
\showDOI{\tempurl}


\bibitem[Fratczak et~al\mbox{.}(2021)]%
        {Fratczak2021RobotInteraction}
\bibfield{author}{\bibinfo{person}{Piotr Fratczak}, \bibinfo{person}{Yee~Mey
  Goh}, \bibinfo{person}{Peter Kinnell}, \bibinfo{person}{Laura Justham}, {and}
  \bibinfo{person}{Andrea Soltoggio}.} \bibinfo{year}{2021}\natexlab{}.
\newblock \showarticletitle{{Robot apology as a post-accident trust-recovery
  control strategy in industrial human-robot interaction}}.
\newblock \bibinfo{journal}{\emph{International Journal of Industrial
  Ergonomics}}  \bibinfo{volume}{82} (\bibinfo{year}{2021}),
  \bibinfo{pages}{103078}.
\newblock
\showISSN{18728219}
\urldef\tempurl%
\url{https://doi.org/10.1016/j.ergon.2020.103078}
\showDOI{\tempurl}


\bibitem[Gavrilova et~al\mbox{.}(2019)]%
        {gavrilova_pilot_2019}
\bibfield{author}{\bibinfo{person}{Liliya Gavrilova}, \bibinfo{person}{Valeri
  Petrov}, \bibinfo{person}{Arina Kotik}, \bibinfo{person}{Artur Sagitov},
  \bibinfo{person}{Liliya Khalitova}, {and} \bibinfo{person}{Tatyana Tsoy}.}
  \bibinfo{year}{2019}\natexlab{}.
\newblock \showarticletitle{Pilot {Study} of {Teaching} {English} {Language}
  for {Preschool} {Children} with a {Small}-{Size} {Humanoid} {Robot}
  {Assistant}}. In \bibinfo{booktitle}{\emph{2019 12th {International}
  {Conference} on {Developments} in {eSystems} {Engineering} ({DeSE})}}.
  \bibinfo{pages}{253--260}.
\newblock
\urldef\tempurl%
\url{https://doi.org/10.1109/DeSE.2019.00055}
\showDOI{\tempurl}


\bibitem[Gosling et~al\mbox{.}(2003)]%
        {gosling_very_2003}
\bibfield{author}{\bibinfo{person}{Samuel~D Gosling}, \bibinfo{person}{Peter~J
  Rentfrow}, {and} \bibinfo{person}{William~B Swann}.}
  \bibinfo{year}{2003}\natexlab{}.
\newblock \showarticletitle{A very brief measure of the {Big}-{Five}
  personality domains}.
\newblock \bibinfo{journal}{\emph{Journal of Research in Personality}}
  \bibinfo{volume}{37}, \bibinfo{number}{6} (\bibinfo{date}{Dec.}
  \bibinfo{year}{2003}), \bibinfo{pages}{504--528}.
\newblock
\showISSN{0092-6566}
\urldef\tempurl%
\url{https://doi.org/10.1016/S0092-6566(03)00046-1}
\showDOI{\tempurl}


\bibitem[Healey et~al\mbox{.}(2018)]%
        {healey_editors_2018}
\bibfield{author}{\bibinfo{person}{Patrick G.~T. Healey},
  \bibinfo{person}{Jan~P. de Ruiter}, {and} \bibinfo{person}{Gregory~J.
  Mills}.} \bibinfo{year}{2018}\natexlab{}.
\newblock \showarticletitle{Editors' {Introduction}: {Miscommunication}}.
\newblock \bibinfo{journal}{\emph{Topics in Cognitive Science}}
  \bibinfo{volume}{10}, \bibinfo{number}{2} (\bibinfo{year}{2018}),
  \bibinfo{pages}{264--278}.
\newblock
\urldef\tempurl%
\url{https://doi.org/10.1111/tops.12340}
\showDOI{\tempurl}
\newblock
\shownote{\_eprint:
  https://onlinelibrary.wiley.com/doi/pdf/10.1111/tops.12340}.


\bibitem[Honig and Oron-Gilad(2018)]%
        {honig_understanding_2018}
\bibfield{author}{\bibinfo{person}{Shanee Honig} {and} \bibinfo{person}{Tal
  Oron-Gilad}.} \bibinfo{year}{2018}\natexlab{}.
\newblock \showarticletitle{Understanding and {Resolving} {Failures} in
  {Human}-{Robot} {Interaction}: {Literature} {Review} and {Model}
  {Development}}.
\newblock \bibinfo{journal}{\emph{Frontiers in Psychology}}
  \bibinfo{volume}{9} (\bibinfo{year}{2018}).
\newblock
\showISSN{1664-1078}
\urldef\tempurl%
\url{https://doi.org/10.3389/fpsyg.2018.00861}
\showDOI{\tempurl}


\bibitem[hysts(2022)]%
        {hysts_pytorch_mpiigaze_demo_2022}
\bibfield{author}{\bibinfo{person}{hysts}.} \bibinfo{year}{2022}\natexlab{}.
\newblock \bibinfo{title}{pytorch\_mpiigaze\_demo: {Gaze} estimation using
  {MPIIGaze} and {MPIIFaceGaze}}.
\newblock
\newblock
\urldef\tempurl%
\url{https://github.com/hysts/pytorch_mpiigaze_demo}
\showURL{%
\tempurl}


\bibitem[Jocher(2021)]%
        {jocher_yolov5_2021}
\bibfield{author}{\bibinfo{person}{Glenn Jocher}.}
  \bibinfo{year}{2021}\natexlab{}.
\newblock \bibinfo{title}{yolov5: v6.0 - {YOLOv5n} '{Nano}' models, {Roboflow}
  integration, {TensorFlow} export, {OpenCV} {DNN} support}.
\newblock
\newblock
\urldef\tempurl%
\url{https://doi.org/10.5281/zenodo.5563715}
\showDOI{\tempurl}


\bibitem[Jording et~al\mbox{.}(2018)]%
        {jording_social_2018}
\bibfield{author}{\bibinfo{person}{Mathis Jording}, \bibinfo{person}{Arne
  Hartz}, \bibinfo{person}{Gary Bente}, \bibinfo{person}{Martin
  Schulte-Rüther}, {and} \bibinfo{person}{Kai Vogeley}.}
  \bibinfo{year}{2018}\natexlab{}.
\newblock \showarticletitle{The “{Social} {Gaze} {Space}”: {A} {Taxonomy}
  for {Gaze}-{Based} {Communication} in {Triadic} {Interactions}}.
\newblock \bibinfo{journal}{\emph{Frontiers in Psychology}}
  \bibinfo{volume}{9} (\bibinfo{year}{2018}).
\newblock
\showISSN{1664-1078}
\urldef\tempurl%
\url{https://doi.org/10.3389/fpsyg.2018.00226}
\showDOI{\tempurl}


\bibitem[Kontogiorgos et~al\mbox{.}(2020)]%
        {kontogiorgos_embodiment_2020}
\bibfield{author}{\bibinfo{person}{Dimosthenis Kontogiorgos},
  \bibinfo{person}{Sanne van Waveren}, \bibinfo{person}{Olle Wallberg},
  \bibinfo{person}{Andre Pereira}, \bibinfo{person}{Iolanda Leite}, {and}
  \bibinfo{person}{Joakim Gustafson}.} \bibinfo{year}{2020}\natexlab{}.
\newblock \showarticletitle{Embodiment {Effects} in {Interactions} with
  {Failing} {Robots}}. In \bibinfo{booktitle}{\emph{Proceedings of the 2020
  {CHI} {Conference} on {Human} {Factors} in {Computing} {Systems}}}
  \emph{(\bibinfo{series}{{CHI} '20})}. \bibinfo{publisher}{Association for
  Computing Machinery}, \bibinfo{address}{New York, NY, USA},
  \bibinfo{pages}{1--14}.
\newblock
\showISBNx{978-1-4503-6708-0}
\urldef\tempurl%
\url{https://doi.org/10.1145/3313831.3376372}
\showDOI{\tempurl}
\newblock
\shownote{event-place: Honolulu, HI, USA}.


\bibitem[Lee et~al\mbox{.}(2010)]%
        {Lee2010GracefullyServices}
\bibfield{author}{\bibinfo{person}{Min~Kyung Lee}, \bibinfo{person}{Sara
  Kielser}, \bibinfo{person}{Jodi Forlizzi}, \bibinfo{person}{Siddhartha
  Srinivasa}, {and} \bibinfo{person}{Paul Rybski}.}
  \bibinfo{year}{2010}\natexlab{}.
\newblock \showarticletitle{Gracefully Mitigating Breakdowns in Robotic
  Services}. In \bibinfo{booktitle}{\emph{Proceedings of the 5th ACM/IEEE
  International Conference on Human-Robot Interaction}}
  \emph{(\bibinfo{series}{HRI '10})}. \bibinfo{publisher}{IEEE Press},
  \bibinfo{pages}{203–210}.
\newblock
\showISBNx{9781424448937}
\urldef\tempurl%
\url{https://doi.org/10.5555/1734454.1734544}
\showDOI{\tempurl}


\bibitem[Lohan et~al\mbox{.}(2012)]%
        {lohan_tutor_2012}
\bibfield{author}{\bibinfo{person}{Katrin~S. Lohan},
  \bibinfo{person}{Katharina~J. Rohlfing}, \bibinfo{person}{Karola Pitsch},
  \bibinfo{person}{Joe Saunders}, \bibinfo{person}{Hagen Lehmann},
  \bibinfo{person}{Chrystopher~L. Nehaniv}, \bibinfo{person}{Kerstin Fischer},
  {and} \bibinfo{person}{Britta Wrede}.} \bibinfo{year}{2012}\natexlab{}.
\newblock \showarticletitle{Tutor {Spotter}: {Proposing} a {Feature} {Set} and
  {Evaluating} {It} in a {Robotic} {System}}.
\newblock \bibinfo{journal}{\emph{International Journal of Social Robotics}}
  \bibinfo{volume}{4}, \bibinfo{number}{2} (\bibinfo{date}{April}
  \bibinfo{year}{2012}), \bibinfo{pages}{131--146}.
\newblock
\showISSN{1875-4805}
\urldef\tempurl%
\url{https://doi.org/10.1007/s12369-011-0125-8}
\showDOI{\tempurl}


\bibitem[Manrique and Enfield(2015)]%
        {manrique_suspending_2015}
\bibfield{author}{\bibinfo{person}{Elizabeth Manrique} {and}
  \bibinfo{person}{N. Enfield}.} \bibinfo{year}{2015}\natexlab{}.
\newblock \showarticletitle{Suspending the next turn as a form of repair
  initiation: evidence from {Argentine} {Sign} {Language}}.
\newblock \bibinfo{journal}{\emph{Frontiers in Psychology}}
  \bibinfo{volume}{6} (\bibinfo{year}{2015}).
\newblock
\showISSN{1664-1078}
\urldef\tempurl%
\url{https://doi.org/10.3389/fpsyg.2015.01326}
\showDOI{\tempurl}


\bibitem[McCabe and Healey(2018)]%
        {mccabe_miscommunication_2018}
\bibfield{author}{\bibinfo{person}{Rose McCabe} {and} \bibinfo{person}{Patrick
  G.~T. Healey}.} \bibinfo{year}{2018}\natexlab{}.
\newblock \showarticletitle{Miscommunication in {Doctor}–{Patient}
  {Communication}}.
\newblock \bibinfo{journal}{\emph{Topics in Cognitive Science}}
  \bibinfo{volume}{10}, \bibinfo{number}{2} (\bibinfo{year}{2018}),
  \bibinfo{pages}{409--424}.
\newblock
\urldef\tempurl%
\url{https://doi.org/10.1111/tops.12337}
\showDOI{\tempurl}
\newblock
\shownote{\_eprint:
  https://onlinelibrary.wiley.com/doi/pdf/10.1111/tops.12337}.


\bibitem[Metta et~al\mbox{.}(2008)]%
        {Metta2008TheCognition}
\bibfield{author}{\bibinfo{person}{Giorgio Metta}, \bibinfo{person}{Giulio
  Sandini}, \bibinfo{person}{David Vernon}, \bibinfo{person}{Lorenzo Natale},
  {and} \bibinfo{person}{Francesco Nori}.} \bibinfo{year}{2008}\natexlab{}.
\newblock \showarticletitle{The {iCub} humanoid robot: An open platform for
  research in embodied cognition}. In \bibinfo{booktitle}{\emph{PerMIS '08:
  Proceedings of the 8th Workshop on Performance Metrics for Intelligent
  Systems}}. \bibinfo{publisher}{ACM}, \bibinfo{pages}{50--56}.
\newblock
\showISBNx{9781605582931}
\urldef\tempurl%
\url{https://doi.org/10.1145/1774674.1774683}
\showDOI{\tempurl}


\bibitem[Mortensen(2012)]%
        {mortensen_visual_2012}
\bibfield{author}{\bibinfo{person}{Kristian Mortensen}.}
  \bibinfo{year}{2012}\natexlab{}.
\newblock \showarticletitle{Visual initiations of repair–some preliminary
  observations}.
\newblock \bibinfo{journal}{\emph{Challenges and new directions in the
  micro-analysis of social interaction}} (\bibinfo{year}{2012}),
  \bibinfo{pages}{45--50}.
\newblock
\newblock
\shownote{Publisher: Kansai University Division of International Affairs Osaka,
  Japan}.


\bibitem[Pasquale et~al\mbox{.}(2016)]%
        {pasquale_enabling_2016}
\bibfield{author}{\bibinfo{person}{Giulia Pasquale}, \bibinfo{person}{Tanis
  Mar}, \bibinfo{person}{Carlo Ciliberto}, \bibinfo{person}{Lorenzo Rosasco},
  {and} \bibinfo{person}{Lorenzo Natale}.} \bibinfo{year}{2016}\natexlab{}.
\newblock \showarticletitle{Enabling {Depth}-{Driven} {Visual} {Attention} on
  the {iCub} {Humanoid} {Robot}: {Instructions} for {Use} and {New}
  {Perspectives}}.
\newblock \bibinfo{journal}{\emph{Frontiers in Robotics and AI}}
  \bibinfo{volume}{3} (\bibinfo{year}{2016}).
\newblock
\showISSN{2296-9144}
\urldef\tempurl%
\url{https://doi.org/10.3389/frobt.2016.00035}
\showDOI{\tempurl}


\bibitem[Purver et~al\mbox{.}(2018)]%
        {purver_computational_2018}
\bibfield{author}{\bibinfo{person}{Matthew Purver}, \bibinfo{person}{Julian
  Hough}, {and} \bibinfo{person}{Christine Howes}.}
  \bibinfo{year}{2018}\natexlab{}.
\newblock \showarticletitle{Computational {Models} of {Miscommunication}
  {Phenomena}}.
\newblock \bibinfo{journal}{\emph{Topics in Cognitive Science}}
  \bibinfo{volume}{10}, \bibinfo{number}{2} (\bibinfo{year}{2018}),
  \bibinfo{pages}{425--451}.
\newblock
\urldef\tempurl%
\url{https://doi.org/10.1111/tops.12324}
\showDOI{\tempurl}
\newblock
\shownote{\_eprint:
  https://onlinelibrary.wiley.com/doi/pdf/10.1111/tops.12324}.


\bibitem[Schegloff et~al\mbox{.}(1977)]%
        {schegloff_preference_1977}
\bibfield{author}{\bibinfo{person}{Emanuel~A. Schegloff}, \bibinfo{person}{Gail
  Jefferson}, {and} \bibinfo{person}{Harvey Sacks}.}
  \bibinfo{year}{1977}\natexlab{}.
\newblock \showarticletitle{The {Preference} for {Self}-{Correction} in the
  {Organization} of {Repair} in {Conversation}}.
\newblock \bibinfo{journal}{\emph{Language}} \bibinfo{volume}{53},
  \bibinfo{number}{2} (\bibinfo{year}{1977}), \bibinfo{pages}{361--382}.
\newblock
\showISSN{00978507, 15350665}
\newblock
\shownote{Publisher: Linguistic Society of America}.


\bibitem[Tanaka and Matsuzoe(2012)]%
        {tanaka_children_2012}
\bibfield{author}{\bibinfo{person}{Fumihide Tanaka} {and}
  \bibinfo{person}{Shizuko Matsuzoe}.} \bibinfo{year}{2012}\natexlab{}.
\newblock \showarticletitle{Children {Teach} a {Care}-{Receiving} {Robot} to
  {Promote} {Their} {Learning}: {Field} {Experiments} in a {Classroom} for
  {Vocabulary} {Learning}}.
\newblock \bibinfo{journal}{\emph{J. Hum.-Robot Interact.}}
  \bibinfo{volume}{1}, \bibinfo{number}{1} (\bibinfo{date}{July}
  \bibinfo{year}{2012}), \bibinfo{pages}{78--95}.
\newblock
\urldef\tempurl%
\url{https://doi.org/10.5898/JHRI.1.1.Tanaka}
\showDOI{\tempurl}
\newblock
\shownote{Publisher: Journal of Human-Robot Interaction Steering Committee}.


\bibitem[Zhang et~al\mbox{.}(2020)]%
        {zhang_eth-xgaze_2020}
\bibfield{author}{\bibinfo{person}{Xucong Zhang}, \bibinfo{person}{Seonwook
  Park}, \bibinfo{person}{Thabo Beeler}, \bibinfo{person}{Derek Bradley},
  \bibinfo{person}{Siyu Tang}, {and} \bibinfo{person}{Otmar Hilliges}.}
  \bibinfo{year}{2020}\natexlab{}.
\newblock \showarticletitle{{ETH}-{XGaze}: {A} {Large} {Scale} {Dataset} for
  {Gaze} {Estimation} {Under} {Extreme} {Head} {Pose} and {Gaze} {Variation}}.
  In \bibinfo{booktitle}{\emph{Computer {Vision} – {ECCV} 2020}},
  \bibfield{editor}{\bibinfo{person}{Andrea Vedaldi}, \bibinfo{person}{Horst
  Bischof}, \bibinfo{person}{Thomas Brox}, {and} \bibinfo{person}{Jan-Michael
  Frahm}} (Eds.). \bibinfo{publisher}{Springer International Publishing},
  \bibinfo{address}{Cham}, \bibinfo{pages}{365--381}.
\newblock
\showISBNx{978-3-030-58558-7}


\end{thebibliography}

\end{document}